\documentclass[sigconf]{acmart}
\def\BibTeX{{\rm B\kern-.05em{\sc i\kern-.025em b}\kern-.08emT\kern-.1667em\lower.7ex\hbox{E}\kern-.125emX}}
    
\copyrightyear{2019} 
\acmYear{2019} 
\setcopyright{rightsretained} 
\acmConference[ACMSE 2019]{2019 ACM Southeast Conference}{April 18--20, 2019}{Kennesaw, GA, USA}
\acmBooktitle{2019 ACM Southeast Conference (ACMSE 2019), April 18--20, 2019, Kennesaw, GA, USA}
\acmDOI{10.1145/3299815.3314478}
\acmISBN{978-1-4503-6251-1/19/04}

\def\BibTeX{{\rm B\kern-.05em{\sc i\kern-.025em b}\kern-.08emT\kern-.1667em\lower.7ex\hbox{E}\kern-.125emX}}
    
\usepackage{flushend} 
\usepackage{geometry}
\hypersetup{draft}
\usepackage{balance}
\begin{document}

%
% The "title" command has an optional parameter, allowing the author to define a "short title" to be used in page headers.
% \title{LSTM Based Deep Learning Approach for Repayment Prediction of Peer-to-Peer (P2P) Lending Market}
% \title{Incorporation of the Macroeconomic Factor into LSTM Approach for Repayment Prediction of Peer-to-Peer (P2P) Lending Market}
\title{Developing and Improving Risk Models using Machine-learning Based Algorithms}

%
% The "author" command and its associated commands are used to define the authors and their affiliations.
% Of note is the shared affiliation of the first two authors, and the "authornote" and "authornotemark" commands
% used to denote shared contribution to the research.
\author{Yan Wang}

\orcid{1234-5678-9012}
\affiliation{%
  \institution{Kennesaw State University}
  \city{Kennesaw, GA}
  \state{USA}
}
\email{ywang63@students.kennesaw.edu}

\author{Xuelei Sherry Ni}
\affiliation{%
  \institution{Kennesaw State University}
  \city{Kennesaw, GA}
  \state{USA}
}
\email{sni@kennesaw.edu}

% \affiliation{%

% \affiliation{%
 
% \affiliation{%

% \affiliation{%



%
% By default, the full list of authors will be used in the page headers. Often, this list is too long, and will overlap
% other information printed in the page headers. This command allows the author to define a more concise list
% of authors' names for this purpose.
\renewcommand{\shortauthors}{Wang and Ni.}

%
% The abstract is a short summary of the work to be presented in the article.
\begin{abstract}
The objective of this study is to develop a good risk model for classifying business delinquency by simultaneously exploring several machine learning based methods including regularization, hyper-parameter optimization, and model ensembling algorithms.  
The rationale under the analyses is firstly to obtain good base binary classifiers (include Logistic Regression ($LR$), K-Nearest Neighbors ($KNN$), Decision Tree ($DT$), and Artificial Neural Networks ($ANN$)) via regularization and appropriate settings of hyper-parameters. 
Then two model ensembling algorithms including bagging and boosting are performed on the good base classifiers for further model improvement. 
The models are evaluated using accuracy, Area Under the Receiver Operating Characteristic Curve (AUC of ROC), recall, and F1 score via repeating 10-fold cross-validation 10 times. 
The results show the optimal base classifiers along with the hyper-parameter settings are $LR$ without regularization, $KNN$ by using 9 nearest neighbors, $DT$ by setting the maximum level of the tree to be 7, and $ANN$ with three hidden layers. 
Bagging on $KNN$ with $K$ valued 9 is the optimal model we can get for risk classification as it reaches the average accuracy, AUC, recall, and F1 score valued 0.90, 0.93, 0.82, and 0.89, respectively.
\end{abstract}

%
% Keywords. The author(s) should pick words that accurately describe the work being
% presented. Separate the keywords with commas.

%
% A "teaser" image appears between the author and affiliation information and the body 
% of the document, and typically spans the page. 
%   \Description{Enjoying the baseball game from the third-base seats. Ichiro Suzuki preparing to bat.}

%
% This command processes the author and affiliation and title information and builds
% the first part of the formatted document.
\maketitle
\section{Problem and Motivation}
Many studies have demonstrated that the performance of risk models can be improved by using many machine-learning based methods including regularization, hyper-parameter optimization, and ensembling algorithms \cite{ozcift2011classifier}. 
% Many studies have demonstrated that the performance of risk models can be improved by using many machine-learning based methods including regularization, hyper-parameter optimization, and model ensembling algorithms \cite{abbey2001human} \cite{huang2007credit} \cite{zhou2018deep} \cite{ozcift2011classifier}. 
In this study, we aim to develop a good risk model for classifying business delinquency by jointly and comprehensively exploring the effect of the above-mentioned model-improving algorithms. 

\section{Background and Related Work}
Logistic Regression ($LR$) is a widely used technique for binary classification because of its strong interpretability and competitive performance. 
Moreover, regularized $LR$, which leads to a substantial decrease in variance and prediction error, 
outperforms $LR$ in some studies. 
Two commonly used versions of $LR$ include $L1$-regularized $LR$ and $L2$-regularized $LR$, with the former version penalizes the $L1$ norm of the coefficients while the latter version penalizes the $L2$ norm of the coefficients \cite{fort2004classification}. 
Decision Tree ($DT$) is another widely acceptable binary classification technique but different settings of its hyper-parameters could largely affect its performance \cite{hazan2017hyperparameter}.
Similarly, in an Artificial Neural Network ($ANN$) and k-Nearest Neighbor ($KNN$), careful tuning of the hyper-parameters can improve their performance \cite{wang2006neighborhood} \cite{wang2018two}. 
Moreover, model ensembling is another effective way to improve the model performance \cite{wang2019ensemble} \cite{zhang2012ensemble}. 

\begin{figure*}
  \includegraphics[width=\textwidth]{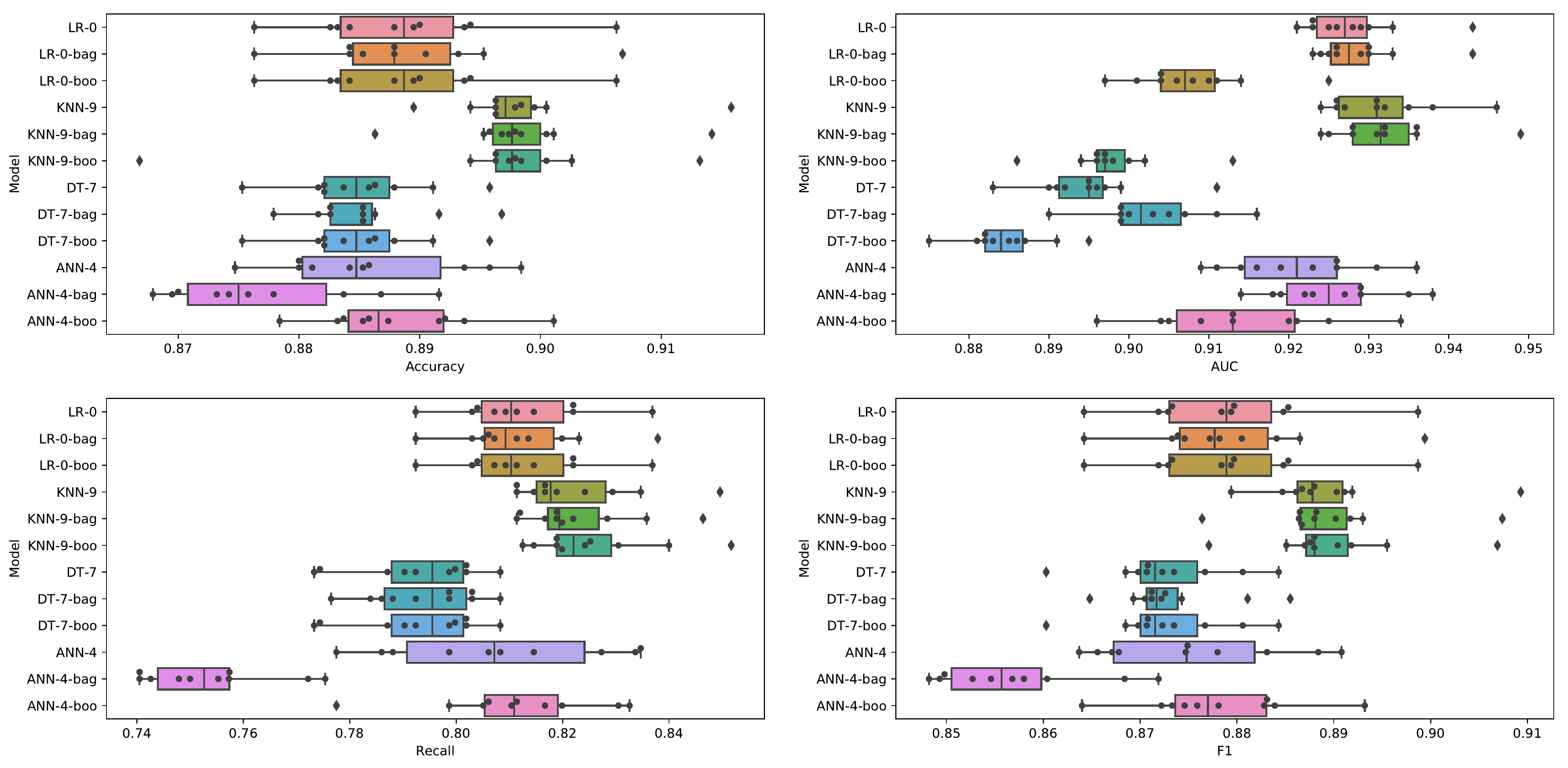}
  \caption{Average Performance of the Best Base Classifiers and their Ensemble Models from 10 Times 10-fold Cross Validation}
  \label{final}
\end{figure*}

\section{Approach}
The dataset used in this study contains the financial information of $9500$ US companies in 2014 with the delinquent rate valued $49.69\%$.
We randomly select $80\%$ of the data as the training set and use the rest as the testing set. 
The dimensionality of the features is reduced to 100 via hierarchical variable clustering in the data pre-processing stage. 
Four base classifiers including $KNN$, $LR$, $DT$ and $ANN$ are developed by using regularization and hyper-parameter optimization algorithms.
To be specific, $LR$ is regularized by using both $L1$ and $L2$-regularization. 
In $KNN$, the hyper-parameter `K', denoting the number of nearest neighbors, is tuned by taking a series of values ranging from 3 to 13. 
In $DT$, we tuned `max\_depth', which denotes the maximum level of the tree structure, by using different values ranging from 5 to 15. 
In $ANN$, the hyper-parameter `layer\_size', representing the number of hidden layers with units in each layer, is tuned by taking a series of values of `50', `50\_25', `50\_25\_13', and `50\_25\_13\_6'.
For example, the value of `50\_25\_13' means there are three hidden layers in the $ANN$ and the number of units in each layer is 50, 25, and 13, respectively. 
Area Under the Receiver Operating Characteristic Curve (AUC of ROC), recall, and F1 score are used as evaluation metrics based on testing set by repeating 10-fold cross validation 10 times \cite{hossin2015review} \cite{saito2015precision}. 
% For the model comparison, Area Under the Receiver Operating Characteristic Curve (AUC of ROC), recall, and F1 score are used as evaluation metrics obtained from the testing set by repeating 10-fold cross validation 10 times \cite{hossin2015review} \cite{goutte2005probabilistic}. 
After building the base classifiers as accurate as possible, two model ensembling techniques (including bagging and boosting) are performed to examine whether the performance can be further improved or not. 

\section{Results}
The best base classifiers along with their hyper-parameter setting are $LR$-0 (i.e, $LR$ without regularization), $KNN$-9 (i.e., $K$ valued 9), $DT$-7 (i.e., `max\_depth' valued 7), and $ANN$-4 (i.e., contains three hidden layers and having units valued 50, 25, and 13 respectively in each layer). 
Figure \ref{final} shows the result of the best classifiers as well as their ensemble models. 
The post-fix bag and boo denote bagging and boosting, respectively. 
In $LR$, bagging and boosting cannot significantly improve the model performance with respect to accuracy, recall, and F1 score. 
Quite unexpectedly, boosting on $LR$ even hurt the AUC by a large extent. 
In $KNN$, both bagging and boosting are beneficial when considering accuracy, recall, and F1 score measures. 
On the contrary, boosting on $KNN$ decrease AUC significantly. 
Bagging on $DT$ outperforms base $DT$ as it produces significantly higher AUC and marginally higher accuracy. 
However, boosting significantly decrease AUC of $DT$. 
Compared with base $ANN$, boosting on $ANN$ is beneficial in terms of accuracy, recall, and F1 score while it hurt AUC significantly. 
Bagging on $ANN$ can significantly decrease accuracy, recall, and F1 score, indicating that $ANN$ is not a good base classifier to be bagged on. 
By comparing all the aforementioned results, we conclude that $KNN$-9-bag (i.e., bagging on the base classifier $KNN$ with $K$ valued 9) is the optimal risk model in our study with average accuracy, AUC, recall, and F1 score valued 0.90, 0.93, 0.82, and 0.89, respectively. 

\section{Conclusions and contributions}
In this paper, we aim at developing and improving the risk modeling via the widely used machine-learning based algorithms including regularization, hyper-parameter optimization, and ensembling simultaneously. 
The results show that the optimal hyper-parameter settings for the base classifiers are $LR$ without regularization, $KNN$ by using 9 nearest neighbors, $DT$ by setting the level of the tree to be 7, and $ANN$ with three hidden layers. 
The optimal model we get for classifying business delinquency is through bagging on $KNN$ with $K$ valued 9, which reaches the average accuracy, AUC, recall, and F1 score valued 0.90, 0.93, 0.82, and 0.89, respectively. 
Although different conclusions may be obtained because of various dataset used in the future, the study methodology provided by us is a good reference for studies that aiming to improve risk modeling using machine-learning based algorithms.

% % The next two lines define the bibliography style to be used, and the bibliography file.
\bibliographystyle{ACM-Reference-Format}
% %\bibliographystyle{ACM BibTeX style}
\bibliography{sample-base}

% 
% If your work has an appendix, this is the place to put it.
% appendix

\end{document}